\pdfoutput=1
\documentclass[11pt]{article}

\usepackage{acl}

\usepackage{times}
\usepackage{latexsym}
\usepackage{hyperref}

\usepackage[T1]{fontenc}
\usepackage[utf8]{inputenc}

\usepackage{microtype}

\usepackage[utf8]{inputenc}
\usepackage{amssymb}
\usepackage{amsmath}
\usepackage{amsthm}
\newtheoremstyle{mystyle}%                % Name
  {}%                                     % Space above
  {}%                                     % Space below
  {\itshape}%                                     % Body font
  {}%                                     % Indent amount
  {\bfseries}%                            % Theorem head font
  {.}%                                    % Punctuation after theorem head
  { }%                                    % Space after theorem head, ' ', or \newline
  {}%                                     % Theorem head spec (can be left empty, meaning `normal')
\theoremstyle{mystyle}

\usepackage{xcolor}

\usepackage{relsize}
\usepackage[Algorithm]{algorithm}
\usepackage{mathtools}
\usepackage{booktabs}
\usepackage{subcaption}
\usepackage{ upgreek }
\usepackage{verbatim} 
\title{TaskDiff: A Similarity Metric for Task-Oriented Conversations}

\author{Ankita Bhaumik$^\dagger$,
    Praveen Venkateswaran$^*$,
    Yara Rizk$^*$,
    Vatche Isahagian$^*$ \\
    $^\dagger$ Rensselaer Polytechnic Institute, Troy, New York \\
    $^*$ IBM Research\\
  \texttt{bhauma@rpi.edu}\\
  \texttt{\{praveen.venkateswaran, yara.rizk, vatchei\}@ibm.com}}

\begin{document}
\maketitle
\begin{abstract}

% April 14
The popularity of conversational digital assistants has resulted in the availability of large amounts of conversational data which can be utilized for improved user experience and personalized response generation. Building these assistants using popular large language models like ChatGPT also require additional emphasis on prompt engineering and evaluation methods. Textual similarity metrics are a key ingredient for such analysis and evaluations.
While many similarity metrics have been proposed in the literature, they have not proven effective for task-oriented conversations as they do not take advantage of unique conversational features. 
To address this gap, we present \textit{TaskDiff}, a novel conversational similarity metric that utilizes different dialogue components (utterances, intents, and slots) and their distributions to compute similarity. Extensive experimental evaluation of \textit{TaskDiff} on a benchmark dataset demonstrates its superior performance and improved robustness over other related approaches. 

\end{abstract}

\section{Introduction}
Task-oriented conversational assistants have become increasingly popular in multiple industries enabling users to perform tasks such as travel reservations, banking transactions, online shopping, etc., through multi-turn conversations.
The increased use of these assistants has led to the availability of valuable user-assistant conversation logs \citep{budzianowski2018multiwoz,andreas2020task}, prompting efforts to extract insights from them. 

A key aspect of such conversational analytics is identifying similarities and dissimilarities between conversations. This will enable developers to improve the user-experience including personalized response generation, next-action recommendations, and information retrieval \citep{yaeli2022recommending,bag2019efficient, gao2020recent,li2022self}.
The popularity of large language models like ChatGPT and Llama 2 \cite{touvron2023llama} has resulted in a race to create custom task-oriented conversational assistants in enterprise domains like finance and retail \cite{wu2023bloomberggpt}. However, evaluating these assistants has become an important challenge and requires effective metrics that can measure their performance across similar user-assistant conversations.

% Existing work on similarity at a high level
Measuring semantic textual similarity has been extensively studied for textual sources like documents, social media, transcripts, etc.  However, there has been limited prior work studying similarity in task-oriented conversation settings \citep{appel2018combining,lavi2021we}.
% Measuring semantic textual similarity is the basis of many natural language and text processing tasks, such as question answering, sentiment analysis, and information extraction \citep{zhou2015learning, ye2016word, poria2016sentic}; it has been extensively studied for textual sources like documents, social media, transcripts, etc. 
% However, there has been limited prior work studying similarity in task-oriented conversation settings \citep{appel2018combining,lavi2021we}.
Most approaches leverage popular word embeddings like Word2Vec \citep{mikolov2013efficient}, GloVe \citep{pennington2014glove} or 
pre-trained models like Universal Sentence Encoder \citep{cer2018universal}, Sentence-BERT \citep{reimers-2019-sentence-bert} to obtain vector representations of utterances, and then use distance-based approaches such as cosine and edit-distance to compute the similarity between text snippets. 

\begin{figure*}[!t]
\centering
    \includegraphics[width=1.0\linewidth]{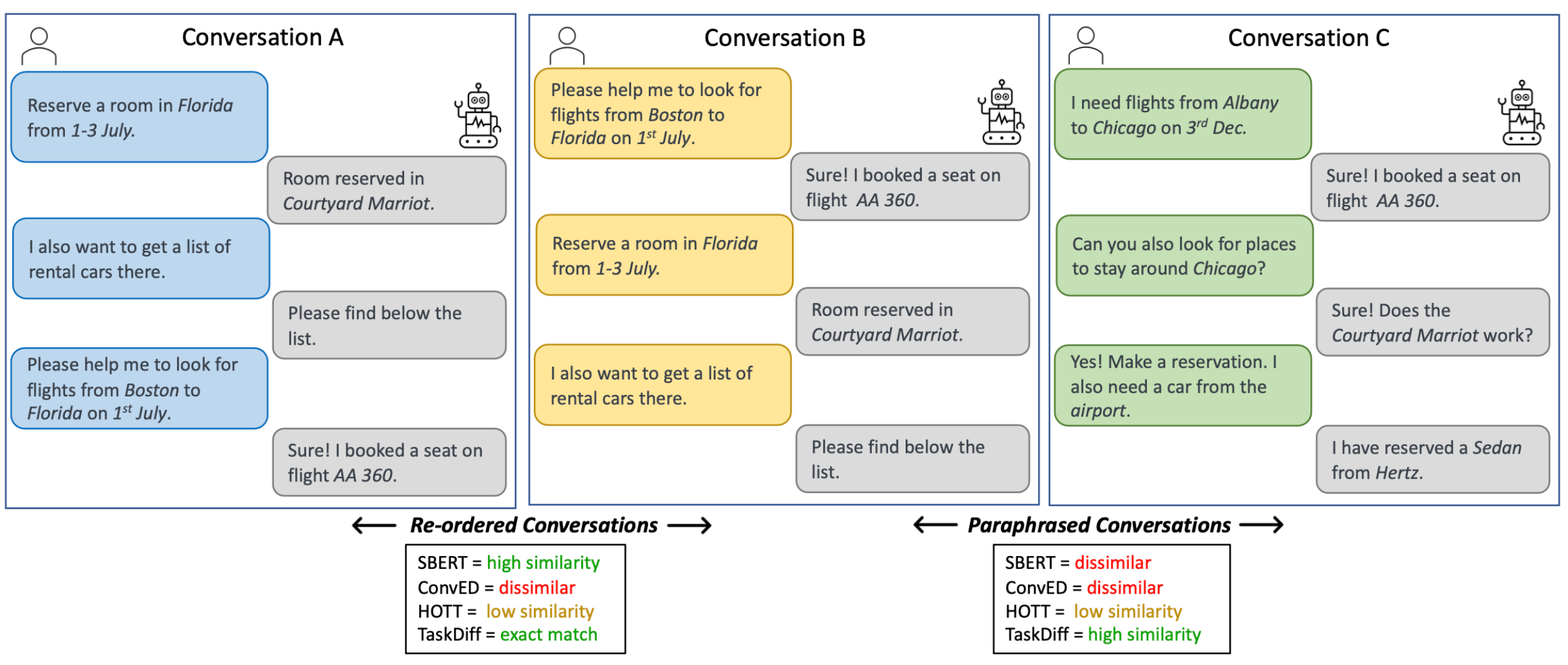}
    \caption{Demonstrating robustness of \textit{TaskDiff} over prior approaches for multiple conversational scenarios.}
    \label{fig:deep_dive}
    % \vspace{-0.2in}
\end{figure*}

While such approaches can identify semantic relationships between texts, task-oriented conversations present several challenges that limit their effectiveness. 
Firstly, they consist of distinct components -- intents, slots, and utterances -- that impact the similarity and overlap between conversations. 
For instance, users can have different objectives (e.g., booking travel vs. product returns), or even have the same intents but provide different levels of slot information \citep{ruane2018botest}. 
Additionally, information is typically provided over multiple conversation turns, and each turn could involve multiple user intents and slots. 
Finally, the same set of tasks can be expressed using numerous possible utterances by users, depending on their choice of phrasing, order of sentences, use of colloquialisms, introducing digressions, etc. \citep{guichard2019assessing}. 
Hence, relying solely on distance based similarity of utterance embeddings would adversely impact performance. 
% For instance, the same utterances of intents and slots when provided in a different order, would result in a low similarity even though the tasks are exactly the same. 

In this work, we present \textit{TaskDiff}, a novel similarity metric designed for task-oriented conversations to address the above challenges. 
Figure \ref{fig:deep_dive} shows multiple users having similar conversations about making bookings for a trip but with re-ordered tasks or paraphrased utterances with different slot values. It also shows that prior work is not robust to such differences that commonly occur in task-oriented conversations.

An ideal metric to measure conversational similarity should be able to identify that the overall goal of these conversations in Figure \ref{fig:deep_dive} is the same.  
\textit{TaskDiff} represents the structure of conversations as distributions over the different task-oriented components and combines the geometry of the distributions with optimal transport to measure the similarity between conversations. Our approach is inspired by prior work in topic modelling \citep{kusner2015word, yurochkin2019hierarchical} that have shown the effectiveness of comparing the structure of distributions, albeit for different settings. 
We evaluate \textit{TaskDiff} on a benchmark task-oriented conversation dataset and demonstrate its effectiveness while presenting examples illustrating its improvement over existing approaches.

\begin{figure*}[!t]
\centering
    \includegraphics[width=\linewidth]{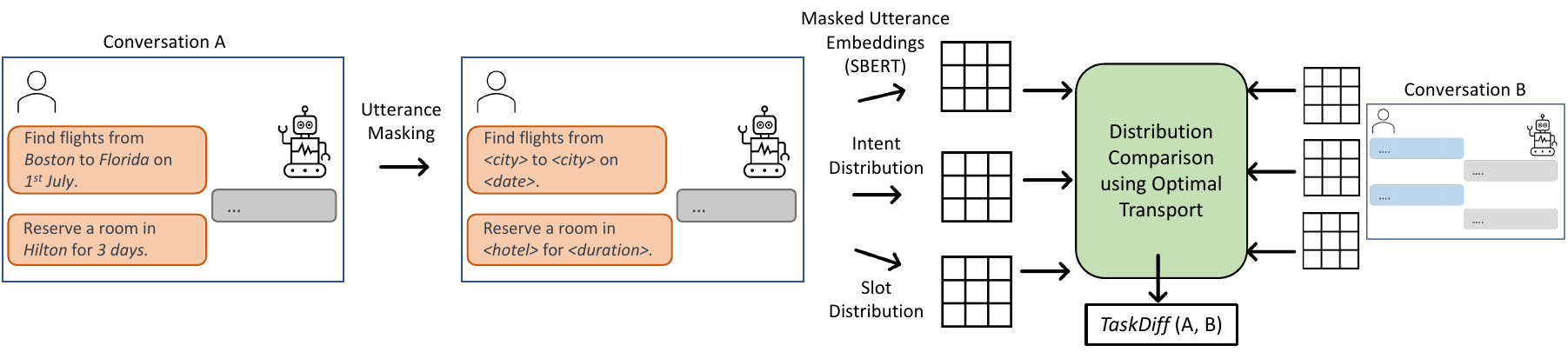}
    \caption{Overview of \textit{TaskDiff} illustrating steps for masking utterances, generating distributions over different conversation components, and computing similarity using Optimal Transport cost between conversations' distributions.}
    \label{fig:TaskDiff_arch}
    % \vspace{-0.1in}
\end{figure*}

% \section{Similarity Metric for Task-Oriented Conversations}
\section{Task-Oriented Conversation Similarity}

\subsection{Definitions}
A task-oriented conversational system supports a pre-defined set of user intents $\mathcal{I}$ and their corresponding slots or parameters $\mathcal{S}$. 
Each conversation $C_i$ consists of a multi-turn sequence of utterances $U_i$  between the user and the system or agent, a subset of active intents, and slot-value information provided by the user (i.e.) $I_i \subseteq \mathcal{I}$ and $S_i \subseteq \mathcal{S}$.
Our objective is to compute the similarity between task-oriented conversations, given their components $K = \{U, \mathcal{I}, \mathcal{S}\}$ (i.e.) utterances, intents, and slot information. 

\subsection{Approach}\label{sec:approach}

% Steps:\\
% Mask slot values in the utterance.\\
% Get intent distribution in the conversation.\\
% Get padded distribution for a pair of conversations.\\
% Get cost matrix as pairwise distance between set of intents.\\
% Get optimal transport distance between the distributions.\\
% Repeat same for slot descriptions and utterances.\\
% Get weighted average of all three distances.\\

\textit{TaskDiff} measures similarity between task-oriented conversations as a function of the distance between their component-wise distributions. 
For each component $k \in K$, we represent its distribution over every conversation and compute similarity as the cumulative cost of transforming or transporting the component-wise distributions of one conversation to another.

Figure \ref{fig:TaskDiff_arch} shows an overview of \textit{TaskDiff}. We first mask the values of the slots in every conversation with their corresponding `<slot name>' from the ontology, before using SBERT to generate conversational embeddings. 
The masking ensures that entities representing the slot values do not incorrectly bias or ambiguate the embeddings \citep{shi2018word}.
For instance, the embedding similarity between the unrelated utterances - \textit{"I want a ticket to the Big Apple"} and \textit{"I want a ticket to the Apple conference"}, could be incorrectly influenced by the word `Apple', but masking with their appropriate slot names (e.g., <arrival$\_$city> and <product$\_$name>), resolves this possibility. We denote $\Delta^l_U$ as the distribution of utterance embeddings of a single conversation.

We then compute probability distributions $\Delta^n_{\mathcal{I}}$, $\Delta^m_{\mathcal{S}}$ for each conversation over the set of intents $\mathcal{I}$ and slots $\mathcal{S}$ as --
    \begin{equation*}
\Delta^n_K = \{ p_i \in \mathbb{R}^{n+1} \;| \sum_{i=0}^n p_i = 1 \;,\; p_i \geq 0 \;\forall i \in |K| \}
\end{equation*}
where each $p_i$ reflects the frequency of occurrence of intents and slots over the utterances. 
For example, $\Delta^n_{\mathcal{I}}$ for conversation $C_i$ represents the probability of all $n$ intents within $C_i$. 

We then compute a separate cost matrix $\mathcal{M}_{i,j}$ for each component, that represents the cost to move between two points ($i,j$) in its distribution. We compute each entry using the Euclidean distance between the embeddings generated for each component. 
% Intuitively, the cost to move across intents/slots closer in the embedding space is lesser than the cost to move across further away intents/slots.
Intuitively, conversations with similar intents, slot information, and analogous language would reflect similar distributions, and hence a lower cost of transportation (i.e.) high similarity. However, any differences in their components would incur a larger cost, and hence reflect a lower similarity. 

Given distributions $\alpha \in \Delta^a_k$, $\beta \in \Delta^b_k, \; \forall k \in K$ and the cost matrix $\mathcal{M}$, the 1-Wasserstein optimal transport distance \citep{vallender1974calculation} between them is --
% The final step for each component is to determine the optimal plan to transport one discrete conversation distribution to the other. The distance between them can be calculated using 1-Wasserstein distance on the conversation distributions and cost matrix. The distance between two distributions $I$ and $J$ can be calculated as:
\begin{gather*}
    W_1(p , q) = \min_{\Gamma \in \mathbb{R}^{n \times m}} \sum\nolimits_{i,j} \mathcal{M}_{i,j}\Gamma_{i,j} \\
    \text{subject to }
    \sum\nolimits_{j}\Gamma_{i,j} = \alpha_i \;\text{and}\; \sum\nolimits_{i}\Gamma_{i,j} = \beta_j
\end{gather*}

where $\mathcal{M}_{i,j} = d(i, j)$ denotes the cost matrix and $d(.,.)$ denotes the distance between the distributions. 
We then define the similarity (\textit{TaskDiff}) between two task-oriented conversations $C_1$ and $C_2$ as the weighted sum of the $W_1$ distances between their respective components -- 
\begin{equation}
    \text{TaskDiff}(C_1, C_2) = \sum_{k=1}^{|K|} \gamma_k W_1(C_1^{\oplus}, C_2^{\oplus})
\end{equation}
where $C_i^{\oplus} = \{U_i, I_i, S_i\}$ 
represents the conversation's components $K$ (i.e.) utterances, intents, and slots, and $\gamma_k$ is a hyperparameter reflecting the influence of each component on the similarity. 
% How are the weights chosen?
% In our experiments we have alpha=2, beta=gamma=1

\section{Experimental Evaluation}

\subsection{Dataset}
We use SGD \citep{rastogi2020towards}, a benchmark dataset of multi-turn task-oriented conversations between users and agents spanning 20 domains (e.g., travel, dining). Its 20,000 conversations are annotated with active intents and slot information.

\subsection{Baselines}
We compare \textit{TaskDiff} to three existing approaches: 

\begin{enumerate}
    \item \textbf{SBERT}: A state-of-the-art approach to measure similarity between conversational embeddings using cosine similarity \citep{reimers-2019-sentence-bert}. 
    \item \textbf{Conversational Edit Distance (ConvED)}: A dialogue similarity metric that aligns utterances between conversations and computes the edit distance between their embeddings \citep{lavi2021we}.
    \item \textbf{Hierarchical Optimal Transport (HOTT)}: A document similarity metric that by models topics using Latent Dirichlet Allocation (LDA) \citep{blei2003latent}, and subsequently uses the 1-Wasserstein distance on the topic and text embeddings \citep{yurochkin2019hierarchical}. 
\end{enumerate}
% \noindent \textbf{SBERT}: A state-of-the-art approach to measure similarity between conversational embeddings using cosine similarity \citep{reimers-2019-sentence-bert}. 

% \noindent \textbf{Conversational Edit Distance (ConvED)}: A dialogue similarity metric that aligns utterances between conversations and computes the edit distance between their embeddings \citep{lavi2021we}.

% \noindent \textbf{Hierarchical Optimal Transport (HOTT)}: A document similarity metric that by models topics using Latent Dirichlet Allocation (LDA) \citep{blei2003latent}, and subsequently uses the 1-Wasserstein distance on the topic and text embeddings \citep{yurochkin2019hierarchical}. 

We conduct our experiments on an Intel Core i9 with 64GB of RAM. 
We implement \textit{TaskDiff} in Python, leveraging the POT library \citep{flamary2021pot} for the 1-Wasserstein optimal transport distance. 
% Publicly available code for approaches from the literature were run to compare performance. 
% The comparison approaches were implemented using their available code, and 
The choice of $\gamma$ was set to 2, 1, and 1 for the intent, utterances and slots components, respectively after performing hyper-parameter search.

% \subsection{Results}
% In this subsection, we show results from a set of classification and clustering experiments that highlight the efficacy and efficiency of our approach.

\subsection{$k$-NN Classification}
We evaluate the ability of the different approaches to accurately classify similar SGD conversations into the correct domains using $k$-NN.
From Table \ref{tab:knn}, we observe that \textit{TaskDiff} outperforms SBERT, HOTT and ConvED, demonstrating the importance of considering other conversational components for similarity beyond just utterances (i.e.) intents and slots, and the need for masking to avoid the adverse influence of entities. The utterance alignment coupled with use of edit distance in ConvED helps compared to SBERT, but requires annotations for alignment that may not always be available. 
We also see that HOTT returns the lowest accuracy, since LDA often picks topics outside the actual conversational intents due to its reliance on word-frequencies. This incorrectly skews the optimal transport distributions thereby impacting classification.

\begin{table}[!h]
\centering
{\small
\begin{tabular}{lc}
\hline
 \textbf{Approach} & \textbf{Accuracy}\\
\hline
SBERT & 0.78 \\ 
HOTT & 0.15\\ 
ConvED &  0.86 \\ 
% Cosine Distance & 0.91 \\ 
\textit{TaskDiff} &  \textbf{0.95}\\ 
\hline
\end{tabular}
\caption{\label{tab:knn}
Accuracy scores for $k$-NN classification
}
}
% \vspace{-0.2in}
\end{table}

\subsection{Conversational Clusters}
\begin{figure}[!tb]
    \includegraphics[width=\linewidth]{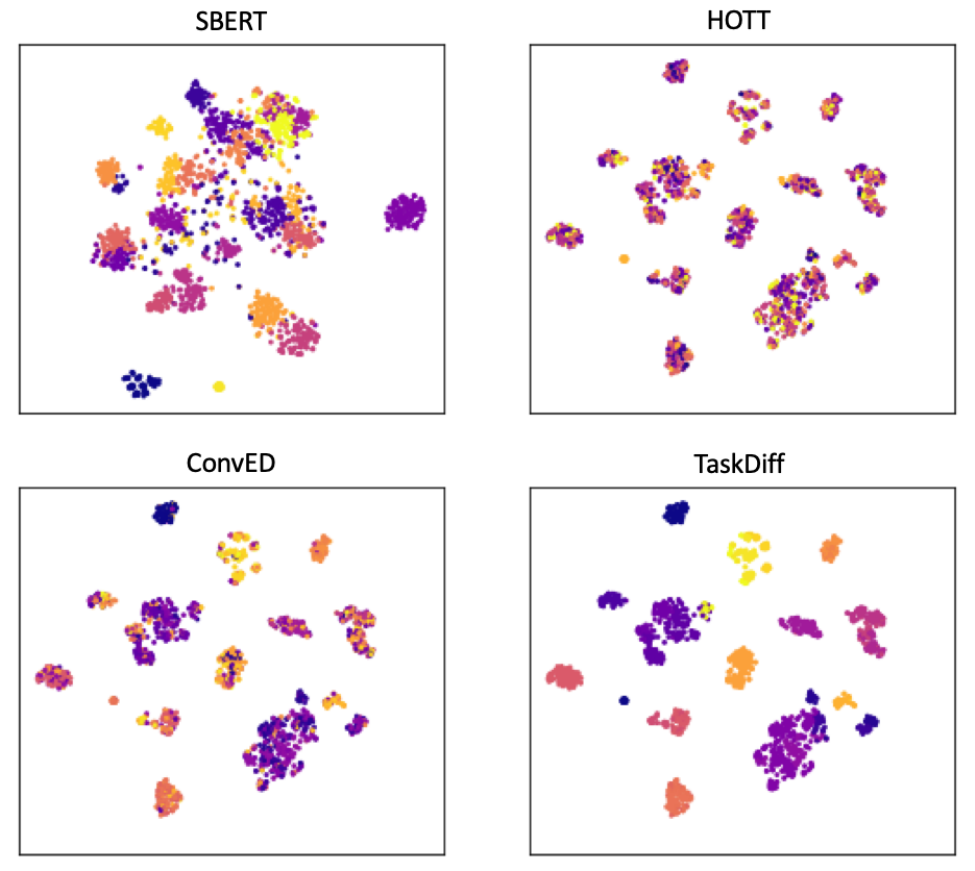}
    \caption{Conversations clustered using $k$-means and color coded by domain names.}
    \label{fig:kmeans_clusters}
    % \vspace{-0.2in}
\end{figure}

We visualize the conversational clusters formed by the different approaches on SGD using $k$-means, setting $k$ to 20 (i.e.) the number of domains and running 20 iterations. From Figure \ref{fig:kmeans_clusters}, we observe that \textit{TaskDiff} results in the most well-formed and distinct clusters followed by SBERT, which has some cluster overlap and lower distinction. The clusters resulting from ConvED and HOTT show a significant amount of overlap, demonstrating their inability to distinguish between similar and dissimilar conversations.

\subsection{Ablation Study}
We perform an ablation study using 200 randomly selected dialogues, to highlight the influence of the different components in \textit{TaskDiff} that enable its effectiveness over approaches like SBERT.
As shown in Table \ref{tab:ablation}, masking the slot names within the utterances results in a $14\%$ improvement in accuracy over SBERT, since the embedding similarity is no longer influenced by incorrect biases or ambiguity from the slot values as described in Section \ref{sec:approach}. 

Additionally, we see that the use of optimal transport (OT) based similarity on the utterances without the use of masks, suffers from the same drawbacks compared to when masks are introduced. Finally, the addition of intents and slots to the optimal transport (i.e.) \textit{TaskDiff} results in a $26\%$ improvement in accuracy over SBERT, due to the additional information about the dialogues provided by these components, thereby highlighting their importance while measuring task-oriented conversation similarity.

\begin{table}[!h]
{\small
\centering
\begin{tabular}{lc}
\hline
\textbf{Approach} & \textbf{Accuracy} \\
\hline
SBERT & 0.73 \\
SBERT + Masking & 0.83 \\
SBERT + OT & 0.68 \\
SBERT + OT + Masking & 0.85 \\
\textit{TaskDiff} & \textbf{0.92} \\

\hline
\end{tabular}
\caption{\label{tab:ablation}
Ablation study of \textit{TaskDiff} with $k$-NN classification accuracy
}
}
% \vspace{-0.2in}
\end{table}

\subsection{Robustness to Reordering}
We evaluate the robustness of the approaches for a common setting where users provide the same tasks in a different order within the conversation. We perturb the SGD dataset, wherein 30\% of the utterances in each conversation are reordered, and compute their distance from the original for each approach. The average distance over all perturbed conversations in Table \ref{tab:re-order} shows that \textit{TaskDiff} returns an exact match on these conversations, since representing conversations as distributions over its components (i.e., intents, slots, utterances), makes it agnostic and robust to such changes. 
The comparison approaches however, are not as robust, with ConvED performing poorly due to its reliance on alignments between utterances.

\begin{table}[!h]
{\small
\centering
\begin{tabular}{lc}
\hline
\textbf{Approach} & \textbf{Avg. Distance} \\
\hline
SBERT & 0.005 \\
HOTT &  0.200\\
ConvED &  4.150\\
\textit{TaskDiff} &  \textbf{0.000}\\
\hline
\end{tabular}
\caption{\label{tab:re-order}
Impact of conversational reordering% on performance of all approaches
}
}
\vspace{-0.2in}
\end{table}

\section{Related Work}\label{sec:related_work}

Efforts across many natural language tasks including sentiment analysis \citep{poria2016sentic}, recommendation systems \citep{magara2018comparative}, and question answering \citep{sidorov2015computing}, have relied on using distance-based similarity measures over text embeddings \citep{wang2020survey}. 
Furthermore, recent work on dialogue similarity have also leveraged conversation structure, where \citet{appel2018combining} consider the number of dialogue turns, words, and cycles and use cosine similarity. Similarly, \citet{xu2019clustering} cluster user-bot dialogues using different distance measures and \citet{enayet2022analysis} measure similarity of dialogue sequences using the Hamming distance.
% and \citet{lavi2021we} align user-bot utterances before using edit-distance to compute similarity. 

The use of optimal transport over text distributions has shown promising results in document similarity \citep{solomon2018optimal} resulting in popular metrics like the word mover's distance (WMD) \citep{kusner2015word} and supervised WMD \citep{huang2016supervised}. Recently, \citet{yurochkin2019hierarchical} used optimal transport over topic models for documents, demonstrating a significant improvement in performance over traditional distance based measures. However, direct application of such approaches to task-oriented dialogues is challenging, due to the unique structure and different components of conversations, as shown in our results.

\section{Conclusion}

In this paper we present \textit{TaskDiff}, a novel metric to measure the similarity between task-oriented conversations. It not only captures semantic similarity between the utterances but also utilizes dialog specific features like intents and slots to identify the overall objective of the conversations. 
We demonstrate that unlike existing metrics, taking advantage of these unique components is critical and results in significantly improved performance. 
As part of future work, we will investigate the inclusion of additional dialog features on open domain dialog datasets and the utilization of \textit{TaskDiff} to improve the performance of various downstream conversational tasks.

\section{Limitations}
We demonstrate in this work that \textit{TaskDiff} is a superior and more robust similarity metric compared to existing state-of-the-art approaches for task-oriented conversations. 
Given the use of optimal transport to compute similarity as a function of differences over the component distributions (intents, slots, and utterances), \textit{TaskDiff} is reliant on being given an ontology for the intents and slots present across the conversations. 
However, this is a fair assumption to make for the domain of task-oriented conversations, and such ontologies are leveraged by real-world deployments such as Google DialogFlow, IBM Watson Assistant, Amazon Lex, etc.

\bibliographystyle{acl_natbib}
\bibliography{references}

% \newpage
% \appendix
% \input{appendix}

\end{document}